\documentclass[a4paper,UKenglish,cleveref, autoref, thm-restate]{lipics-v2021}
\usepackage{amsmath,amsfonts,bm}

\def\eqref#1{equation~\ref{#1}}
\def\1{\bm{1}}

\DeclareMathAlphabet{\mathsfit}{\encodingdefault}{\sfdefault}{m}{sl}
\SetMathAlphabet{\mathsfit}{bold}{\encodingdefault}{\sfdefault}{bx}{n}

\usepackage{microtype}
\usepackage{graphicx}
\usepackage{booktabs} % for professional tables

\usepackage{hyperref}
\usepackage{url}

\usepackage[utf8]{inputenc} % allow utf-8 input
\usepackage[LGR, T1]{fontenc}
\usepackage{hyperref}       % hyperlinks
\usepackage{url}            % simple URL typesetting
\usepackage{booktabs}       % professional-quality tables
\usepackage{amsfonts}       % blackboard math symbols
\usepackage{nicefrac}       % compact symbols for 1/2, etc.
\usepackage{microtype}      % microtypography
\usepackage{xcolor}         % colors
\usepackage{makecell}
\usepackage[framemethod=TikZ]{mdframed}
\usepackage{amssymb}% http://ctan.org/pkg/amssymb
\usepackage{pifont}% http://ctan.org/pkg/pifont
\usepackage{float}

\usepackage[frozencache, cachedir=_minted]{minted}
\usepackage{minted}
\usepackage{wrapfig}
\usepackage{clrscode}
\usepackage{subcaption}
\usepackage{color}
\definecolor{keywordcolor}{rgb}{0.7, 0.1, 0.1}   % red
\definecolor{tacticcolor}{rgb}{0.0, 0.1, 0.6}    % blue
\definecolor{commentcolor}{rgb}{0.4, 0.4, 0.4}   % grey
\definecolor{symbolcolor}{rgb}{0.0, 0.1, 0.6}    % blue
\definecolor{sortcolor}{rgb}{0.1, 0.5, 0.1}      % green
\definecolor{attributecolor}{rgb}{0.7, 0.1, 0.1} % red

\usepackage{listings}

\newtheorem{problem}{Problem}
\renewcommand{\paragraph}[1]{\noindent \textbf{#1}}

\DeclareUnicodeCharacter{2223}{$\mid$}
\DeclareUnicodeCharacter{211D}{$\mathbb{R}$}
\DeclareUnicodeCharacter{2081}{$_{1}$}
\DeclareUnicodeCharacter{2082}{$_{2}$}
\DeclareUnicodeCharacter{22A2}{$\vdash$}
\DeclareUnicodeCharacter{2115}{$\mathbb{N}$}
\DeclareUnicodeCharacter{2211}{$\sum$}
\DeclareUnicodeCharacter{2194}{$\leftrightarrow$}
\DeclareUnicodeCharacter{2208}{$\in$}
\DeclareUnicodeCharacter{1D55C}{$\mathbb{k}$}
\DeclareUnicodeCharacter{3B9}{$\iota$}
\DeclareUnicodeCharacter{3C0}{$\pi$}
\DeclareUnicodeCharacter{271D}{$\dagger$}
\DeclareUnicodeCharacter{2075}{$^{5}$}
\DeclareUnicodeCharacter{2074}{$^{4}$}
\DeclareUnicodeCharacter{2073}{$^{3}$}
\DeclareUnicodeCharacter{2072}{$^{2}$}
\DeclareUnicodeCharacter{2071}{$^{1}$}
\DeclareUnicodeCharacter{2200}{$\forall$}
\DeclareUnicodeCharacter{2A0D}{$\int$}
\DeclareUnicodeCharacter{223C}{\textasciitilde}
\usepackage[textsize=scriptsize]{todonotes}

\newcommand{\proofwala}[0]{\textsc{ProofWala}}
\newcommand{\lean}[0]{\textsc{Lean}}
\newcommand{\multi}[0]{\textsc{Multilingual}}
\newcommand{\coq}[0]{\textsc{Rocq}}
\newcommand{\codeTFive}[0]{\textsc{CodeT5}}
\newcommand{\base}[0]{\textsc{Base}}
\newcommand{\HuggingFace}[0]{\textsc{HuggingFace}\xspace}

\usepackage{xspace}
\newcommand{\name}{\textsc{ProofWala}\xspace}

\title{\name: A Framework for Multilingual Proof Data Synthesis and Theorem-Proving} %TODO Please add
\titlerunning{ProofWala}
\author{Amitayush {Thakur}}{University of Texas, Austin, USA \and \url{https://amit9oct.github.io/aboutme/}}{amitayush@utexas.edu}{https://orcid.org/0009-0005-5622-7831}{}

\author{George {Tsoukalas}}{University of Texas, Austin, USA \and \url{https://georgetsoukalas.github.io/}}{george.tsoukalas@utexas.edu}{}{}

\author{Greg {Durrett}}{University of Texas, Austin, USA \and \url{https://gregdurrett.github.io/}}{gdurrett@cs.utexas.edu}{}{}

\author{Swarat {Chaudhuri}}{University of Texas, Austin, USA \and \url{https://www.cs.utexas.edu/~swarat/}}{swarat@cs.utexas.edu}{}{}

\authorrunning{A. Thakur and G. Tsoukalas and G. Durrett and S. Chaudhuri} %TODO mandatory. First: Use abbreviated first/middle names. Second (only in severe cases): Use first author plus 'et al.'

\Copyright{Amitayush Thakur and George Tsoukalas and Greg Durrett and Swarat Chaudhuri} %TODO mandatory, please use full first names. LIPIcs license is "CC-BY";  http://creativecommons.org/licenses/by/3.0/


 %TODO mandatory; please add comma-separated list of keywords

\category{} %optional, e.g. invited paper

\relatedversion{} %optional, e.g. full version hosted on arXiv, HAL, or other respository/website
\nolinenumbers %uncomment to disable line numbering

\EventEditors{Ekaterina Komendantskaya and Tobias Nipkow}
\EventNoEds{2}
\EventLongTitle{17th International Conference on Interactive Theorem Proving (ITP 2026)}
\EventShortTitle{ITP 2026}
\EventAcronym{ITP}
\EventYear{2026}
\EventDate{July 26--29, 2026}
\EventLocation{Lisbon, Portugal}
\EventLogo{}
\SeriesVolume{382}
\ArticleNo{24}
\begin{document}

\maketitle

\begin{abstract}
Neural approaches to theorem proving require robust infrastructure for interfacing with interactive theorem provers (ITPs), extracting structured proof data, and executing proof search at scale. However, existing tooling is often assistant-specific and oriented toward interactive, file-level execution, making repository-scale analysis and parallel experimentation challenging. We present \name, a multilingual proof engineering framework built around \texttt{itp-interface}, a reusable library for programmatic interaction with ITPs. For \lean~4, we implement a meta-programmed interaction layer that executes inside the elaborator, enabling semantically faithful tactic-level tracing together with declaration- and dependency-level extraction across entire repositories. This design extends beyond traditional REPL-style interaction by supporting project-wide analysis, environment cloning, and pooled execution of proof states. The implementation is robust across \lean~4 versions after 4.15.0 with forward compatibility support. The same interface abstraction supports tactic execution and data extraction for multiple versions of \coq, yielding a unified cross-assistant pipeline.

Built on this infrastructure, \name provides standardized multilingual proof datasets, model training utilities, and parallel proof search algorithms. 
Using the framework, we demonstrate that multilingual training across \lean\ and \coq\ enables cross-lingual and cross-domain transfer. In particular, we observe statistically significant improvements on the largest benchmark (\lean/\textsc{Mathlib}) and in domain adaptation (CategoryTheory), while other settings exhibit consistent but not statistically significant trends in favor of multilingual training.
%Using the framework, we demonstrate that multilingual training across \lean\ and \coq\ enables cross-lingual and cross-domain transfer, with mixed-training models outperforming assistant-specific baselines on the standard prove-at-$k$ metric. 
We open-source the full framework, including the parallel proof search module, the \texttt{itp-interface} library, multilingual datasets, and trained models, providing a scalable foundation for proof mining, neural theorem proving, and cross-assistant experimentation.
\end{abstract}

\section{Introduction}
\newcommand{\leanrepl}{\href{https://github.com/leanprover-community/repl}{\lean{} REPL}\xspace}
\newcommand{\pantograph}{\href{https://github.com/leanprover/Pantograph}{\textsc{Pantograph}}\xspace}
\newcommand{\itpinterfaceurl}{https://github.com/trishullab/itp-interface}
\newcommand{\proofwalaurl}{https://github.com/trishullab/proof-wala}
\newcommand{\hfproofwalamultiurl}{https://huggingface.co/amitayusht/ProofWala-Multilingual}
\newcommand{\hfproofwalaleanurl}{https://huggingface.co/amitayusht/ProofWala-Lean}
\newcommand{\hfproofwalacoqurl}{https://huggingface.co/amitayusht/ProofWala-Coq}

Automated theorem proving has long been considered a grand challenge in artificial intelligence. Recently, deep learning has emerged as a promising approach to this challenge \cite{li2024surveydeeplearningtheorem,yang2024formalmathematicalreasoningnew}. Broadly, neural methods for theorem proving use language models to generate proofs in the language of an \emph{interactive theorem prover} (ITP), e.g., \lean\; \cite{de2015lean}, \coq\; \cite{huet1997coq}, or Isabelle \cite{paulson1994isabelle}. An ITP represents proofs as sequences of simplification steps, or \emph{tactics}, and mechanically checks such proofs for correctness. From this perspective, automated theorem proving amounts to generating a tactic sequence that passes the ITP's checks.

Most neural approaches follow the strategy proposed by Polu et al. \cite{polu2020generative}: train a generative language model (LM) to predict tactics (and their parameters) conditioned on the current proof state, using proof-step data extracted from existing formal repositories, and then wrap the model with a search procedure that conducts proof search (see \Cref{sec:problem-formulation}). 

While neural theorem proving is gaining momentum, the supporting \emph{tooling infrastructure} remains fragmented. Existing pipelines for data extraction and proof interaction are often ITP-specific, rely on brittle, assistant-dependent representations, and provide limited support for scalable execution (e.g., parallel tactic evaluation, cloning proof states, or traversing large dependency graphs). This fragmentation hinders systematic comparisons, slows down proof-engineering workflows, and makes it difficult to exploit potential cross-lingual and cross-domain gains from training on multilingual formal corpora.

In response, we introduce \proofwala\footnote{``Wala'' is a suffix from Indic languages (often spelled “wallah”), meaning ``one who is associated with or provides a particular thing.''}, a multilingual framework for dataset collection, interaction, training, and proof search across ITPs and domains. A central component of \proofwala\ is \texttt{itp-interface}, a reusable library for programmatic interaction with proof assistants. For \lean, we deliberately avoid a \leanrepl\footnote{\href{https://github.com/leanprover-community/repl}{https://github.com/leanprover-community/repl}}-based architecture. While REPL-style interfaces are well suited for interactive, file-by-file tactic execution, they primarily expose sequential tactic traces and provide limited access to declaration metadata or project-wide structure. In particular, extracting declaration-level information (e.g., inductive definitions, constants, namespaces, and imports) and constructing repository-wide dependency graphs is difficult or infeasible through a REPL workflow. Such capabilities are essential for our goals of dependency-aware data extraction, large-scale repository mining, and cloning proof states for parallel execution.

Instead, we implement \lean{} meta-programs that run directly inside the elaborator, enabling semantically faithful tactic-level tracing together with declaration and dependency extraction across an entire project. This design supports environment cloning and pooling, which in turn enables parallel proof search and high-throughput data collection. The resulting implementation is robust across \lean\ releases starting from 4.15.0 via a thin compatibility layer that isolates internal API changes and provides largely automatic forward compatibility. The same interface abstraction supports tactic execution and data extraction for \coq\ (multiple versions), yielding a unified cross-assistant pipeline.

Compared to existing \lean{} tooling such as \pantograph\footnote{\href{https://github.com/leanprover/Pantograph}{https://github.com/leanprover/Pantograph}}\cite{pantograph}, which focuses primarily on interactive orchestration of a single \lean\ instance, \texttt{itp-interface} is designed as a scalable infrastructure layer: it supports repository-level extraction, declaration graph construction, multi-environment execution pools, and distributed parallelism (e.g., via Ray~\cite{moritz2018ray}, a framework for parallel and distributed execution across processes or machines), as well as multilingual (\lean{} and \coq{}) interoperability. These capabilities are central to large-scale proof mining and parallel proof search, and extend beyond typical REPL- or single-environment-based workflows.
% Compared to existing \lean{} tooling such as \pantograph\footnote{\href{https://github.com/leanprover/Pantograph}{https://github.com/leanprover/Pantograph}}\cite{pantograph}, which focuses primarily on interactive orchestration of a single \lean\ instance, \texttt{itp-interface} is designed as a scalable infrastructure layer: it supports repository-level extraction, declaration graph construction, multi-environment execution pools, distributed parallelism (e.g., via Ray\cite{moritz2018ray}), and multilingual (\lean{} and \coq{}) interoperability. These capabilities are central to large-scale proof mining and parallel proof search, and extend beyond typical REPL- or single-environment-based workflows.

Beyond interaction and extraction, \proofwala\ integrates efficient proof search algorithms, including parallelized best-first and beam search, enabled by an \emph{environment pool} abstraction that maintains multiple proof environments initialized to identical frontier states. This pool allows executing candidate tactics concurrently, and integrates naturally with Ray-based distributed execution for parallel proof search and large-scale annotation.

We release an open-source code library, multilingual datasets, and fine-tuned models that facilitate end-to-end proof search in \lean\;4 and \coq. Using \proofwala, we demonstrate that multilingual training can yield positive cross-lingual and cross-domain transfer, improving proof step generation across different assistants and domains.

In summary, our work makes three key contributions:

\paragraph{A Standardized Framework and Engineering Library.}
We propose \proofwala, built around \texttt{itp-interface}, a unified framework for extracting and organizing tactic-level training data and supporting tool development for formal theorem proving in \lean\; and \coq. For \lean, we introduce a meta-programmed instrumentation pipeline that supports (i) tactic-level execution with semantically faithful proof-state tracing, including improved handling of nested \texttt{have} blocks in tactic mode, and (ii) dependency-aware repository traversal for extracting declarations and their dependency relations across entire \lean\ projects. Our \lean\ backend is version-robust across \lean~4 releases 4.15.0+ (largely automatic forward compatibility). For \coq, we support tactic execution and data extraction across multiple versions, enabling a consistent unified data format and prompt-formatting scheme across ITPs and domains. All code is open source: the \name{} framework can be found \href{\proofwalaurl}{\textcolor{blue}{here}}, and \texttt{itp-interface} can be found \href{\itpinterfaceurl}{\textcolor{blue}{here}}. We additionally provide a Streamlit~\cite{streamlit}-based \emph{Lean Declaration Database Explorer}—an interactive web application for data exploration that enables querying extracted SQLite databases, searching declarations, and visualizing dependency structure. All our fine-tuned models \proofwala-\{\lean, \coq, and \multi\} on \HuggingFace for public use (URL: \href{\hfproofwalaleanurl}{\textcolor{blue}{\lean}}, \href{\hfproofwalacoqurl}{\textcolor{blue}{\coq}} and \href{\hfproofwalamultiurl}{\textcolor{blue}{\multi}}).
%We additionally provide a Streamlit\cite{streamlit}-based \emph{Lean Declaration Database Explorer} for interactively querying extracted SQLite databases, searching declarations, and visualizing dependency structure. We will also release all our fine-tuned models \proofwala-\{\lean, \coq, and \multi\} on \HuggingFace

\paragraph{Support for Parallel Proof Completion.}
Similar to \textit{GPT-f} \cite{polu2020generative}, the framework supports proof completion via search guided by a proof step generation model. We make proof search ITP-agnostic and parallel by introducing an \emph{environment pool} that clones proof environments and executes tactics in parallel across these replicas (scaling across machines using Ray). This enables high-throughput exploration, proof-tree annotation, and visualization. To the best of our knowledge, \proofwala\ is the first open-source framework to provide parallel proof search by explicitly supporting cloned proof environments and concurrent tactic evaluation within a unified multi-ITP interface.

% \paragraph{Demonstration of Cross-Lingual and Cross-Domain Transfer.}
% Using \proofwala, we study the effect of multilingual proof data on proof step prediction, and observe cross-domain and cross-lingual transfer for both \lean\; and \coq, across general mathematics and software verification. We release the multi-domain multilingual training data (about 450K datapoints; 270M tokens; about 80K theorems) and multiple open models trained on different data mixtures. Our \proofwala-\multi\; model is an open proof step generation model trained on diverse domains and ITPs, and can be directly used for proof search in both formal mathematics and software verification.

\paragraph{Demonstration of Cross-Lingual and Cross-Domain Transfer.}
Using \proofwala, we study the effect of multilingual proof data on proof step prediction, and observe evidence of cross-domain and cross-lingual transfer for both \lean\; and \coq, across general mathematics and software verification. We observe statistically significant improvements on the largest benchmark (\lean/\textsc{Mathlib}) and in domain adaptation (CategoryTheory), while other settings exhibit consistent but not statistically significant trends in favor of multilingual training.

We release the multi-domain multilingual training data (about 450K datapoints; 270M tokens; about 80K theorems) and multiple open models trained on different data mixtures. Our \proofwala-\multi\; model is an open proof step generation model trained on diverse domains and ITPs, and can be directly used for proof search in both formal mathematics and software verification.

\section{Problem Formulation}
\label{sec:problem-formulation}

\newcommand{\init}{\mathit{in}}
\newcommand{\error}{\mathit{error}}
\newcommand{\leqhard}{\sqsubseteq}
\newcommand{\geqhard}{\sqsupseteq}
\newcommand{\QED}{\mathtt{QED}}
\newcommand{\Ob}{\mathcal{O}}
\newcommand{\Ib}{\mathcal{I}}
\newcommand{\glob}{\chi} 
\renewcommand{\error}{\mathit{Err}}

\newcommand{\edge}[1]{\stackrel{#1}{\longrightarrow}}

We model interactive theorem proving as a structured state-transition system operating over \emph{proof obligations}. Each obligation $o$ is a pair $(g, h)$, where $g$ is the current goal and $h$ is the associated local context (hypotheses). A proof state $O$ is a finite set of such obligations. The objective of the prover is to transform an initial state into a terminal state in which all obligations have been discharged.

Formally, we abstract an ITP as a \emph{proof environment} consisting of:

\begin{itemize}
    \item A set of states $\Ob$, where each state $O = \{o_1, \dots, o_k\}$ is a set of obligations.
    \item An initial state $\Ib = \{(g_{\init}, h_{\init})\}$ extracted from a user-provided theorem.
    \item A distinguished goal state $\QED$, corresponding to the empty obligation set.
    \item A finite (or effectively enumerable) set of tactics.
    \item A transition function $T(O, a)$ describing the result of applying tactic $a$ to state $O$.
\end{itemize}

If tactic $a$ is successfully applied at state $O$, then $T(O,a)$ yields a new set of obligations reflecting the updated proof state. If $a$ is invalid or fails, we treat the transition as leaving the state unchanged.\footnote{In practice, failure is captured explicitly by the underlying ITP; our abstraction omits low-level error states for clarity.}

We extend $T$ to sequences of tactics $\alpha = \langle a_1, \dots, a_n \rangle$ via:

\[
T_{seq}(O, \alpha) =
\begin{cases}
T(O, a_1) & \text{if } n = 1 \\
T(T_{seq}(O, \langle a_1, \dots, a_{n-1} \rangle), a_n) & \text{otherwise.}
\end{cases}
\]

\begin{problem}[Theorem Proving]
Given an initial state $O_{\init}$, find a tactic sequence $\alpha$ such that
\[
T_{seq}(O_{\init}, \alpha) = \QED.
\]
\end{problem}

This abstraction is independent of any specific proof assistant and underlies the design of \texttt{itp-interface}. In our implementation, Lean~4 and Coq proof states are mapped into this unified representation via meta-programmed instrumentation (Lean) and assistant-specific interaction layers (Coq). This uniform state model enables assistant-agnostic data extraction, dependency-aware analysis, environment cloning, and parallel proof search.

\begin{wrapfigure}{r}{3.9in}
\vspace{-0.3in}
\begin{mdframed}[roundcorner=10pt]
\begin{minipage}{0.9\textwidth}
\begin{lstlisting}
theorem blockTriangular_stdBasisMatrix 
{i j : m} (hij : b i ≤ b j) (c : R) 
: BlockTriangular (stdBasisMatrix i j c) b  
:= by rintro i' j' hij'
  simp [stdBasisMatrix, hij, hij'.not_le]
  rintro rfl rfl
  exact (not_lt_of_le hij hij').elim
\end{lstlisting}
\end{minipage}
\end{mdframed}
%\reducevspacebetweenfigureandcaption
\vspace{-0.1in}
\caption{A \lean\; 4 theorem and a with a correct proof using \proofwala-\multi\; proof-step generation model. The theorem states that the standard basis matrix, where $c$ is placed in the $(i,j)th$ entry with zeroes elsewhere is block triangular. The first tactic \texttt{rintro i' j' hij'} unfolds the definition of \texttt{BlockTriangular} and adds the variables \texttt{i'}, \texttt{j'}, as well as the hypothesis \texttt{hij' : b j' < b i'} to the set of hypotheses. The proof proceeds by using established properties of the \texttt{stdBasisMatrix} and resolves by demonstrating an inconsistency with the hypothesis \texttt{hij : b i $\leq$ b j}.}
\label{fig:example}
\vspace{-0.25in}
\end{wrapfigure}

\Cref{fig:example} shows a \lean theorem and proof discovered using our framework. The proof proceeds by successively transforming the initial obligation through tactic applications until the obligation set is empty.

\paragraph{Model-Guided Search.}
To automate the search for $\alpha$, we employ a model that estimates the likelihood of applying tactic $a$ at state $O$, denoted $p(a \mid O)$. In practice, this model may be implemented as a language model that generates tokenized tactic representations conditioned on a serialized form of $O$. We omit tokenization details for clarity. Because our state abstraction is standardized across assistants, models can be trained on multilingual proof data and applied uniformly across \lean{} and \coq{} environments.

The search module operates over the state space defined above, using $p(a \mid O)$ to guide node expansion. Crucially, our environment-pooling abstraction allows multiple frontier states to be evaluated in parallel, enabling scalable exploration of the proof tree across cloned proof environments.

\section{The \proofwala\ Framework}
We now describe the \proofwala\ framework. Our primary motivation is to provide a reusable, language-agnostic infrastructure for interacting with interactive theorem provers (ITPs), extracting structured proof data at scale, and supporting programmable proof search. Rather than focusing solely on model training, we design \proofwala\ as a systems layer that enables repository-wide mining, dependency-aware analysis, environment cloning, and scalable experimentation across assistants.

The framework is organized around three components:

\begin{enumerate}
    \item \textbf{Interface Module} (\texttt{itp-interface}): a reusable library for programmatic interaction with ITPs.
    \item \textbf{Data and Model Module}: infrastructure for extracting proof traces and training proof-step predictors.
    \item \textbf{Parallel Proof Search Module}: scalable exploration of proof states using environment pooling.
\end{enumerate}

\begin{figure*}
    \centering
    \includegraphics[width=\textwidth]{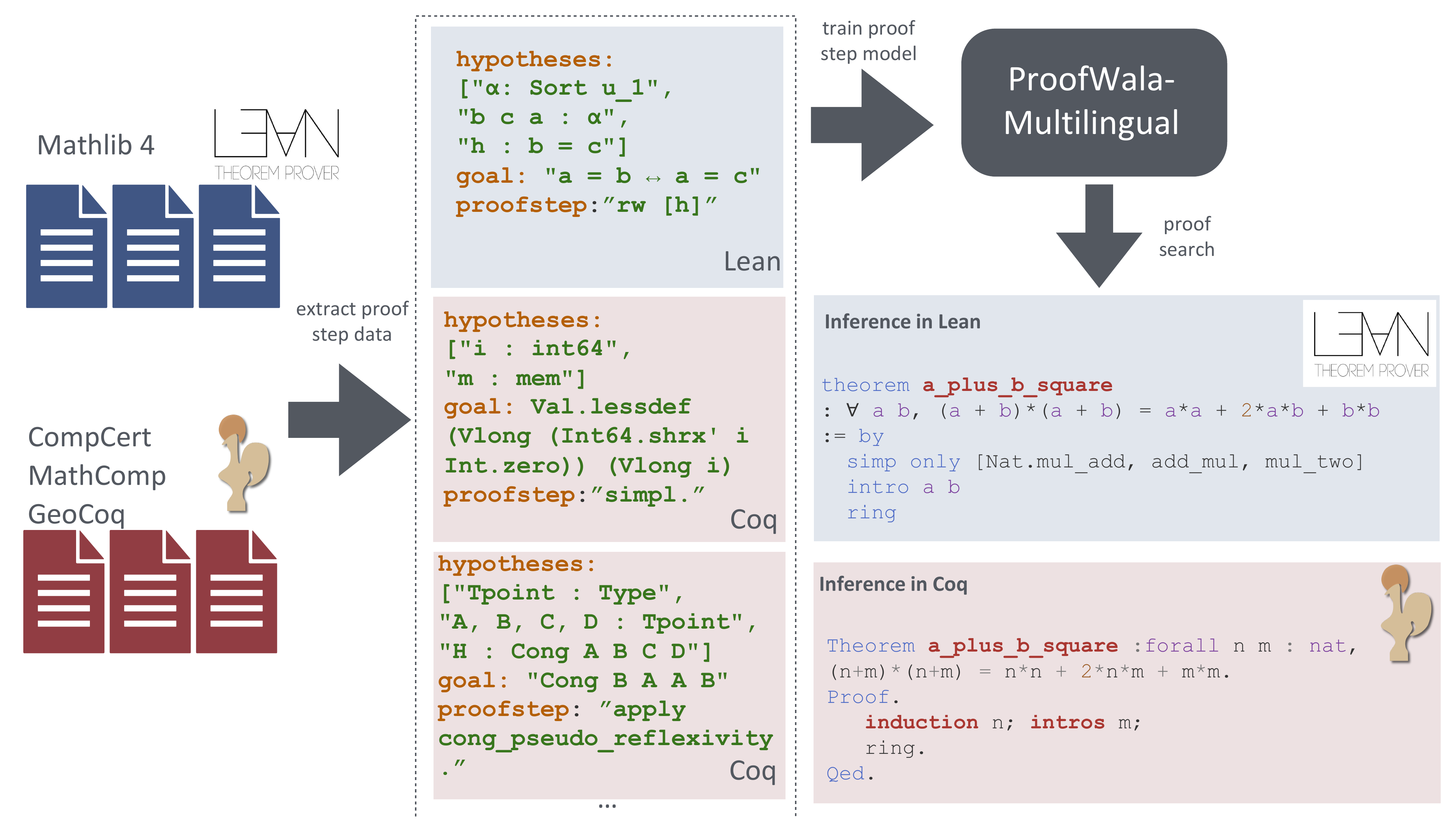}
    \caption{Overview of the \proofwala\ framework. The interface module enables repository-scale extraction and programmable interaction with \lean\; and \coq. Extracted traces are used for training proof-step predictors. The search module executes tactics—potentially in parallel—using cloned proof environments.}
    \label{fig:proofwala-summary}
\end{figure*}

\Cref{fig:proofwala-summary} illustrates how these components interact.

\subsection{Interface Module}
\label{sec:interface-module}

The \texttt{itp-interface} library provides a unified abstraction for interacting with multiple ITPs. In particular, the interface module supports interaction with \lean\;4 and \coq\ (multiple versions from 10.0 to 18.0), and exposes a standardized state-transition API consistent with the formulation in \Cref{sec:problem-formulation}. Despite differences in internal architecture, both assistants are mapped into a common representation of proof states and tactic transitions.

\paragraph{\lean\;4 Backend via Meta-Programming.}
Our \lean\; backend is implemented using \lean\;4 meta-programming rather than relying solely on external orchestration. The interaction layer executes inside the elaborator, enabling:

\begin{itemize}
    \item semantically faithful tactic-level execution and state tracing,
    \item extraction of declaration-level metadata (e.g., inductive definitions, constants, namespaces),
    \item construction of repository-wide dependency graphs,
    \item improved handling of nested tactic constructs (e.g., \texttt{have}, \texttt{by}, and subgoal generation),
    \item cloning of proof environments for scalable parallel execution.
\end{itemize}

This design enables repository-scale mining and dependency-aware extraction that extends beyond file-scoped interaction workflows. The implementation is compatible with \lean\;4 versions 4.15.0 through 4.24.0 and employs a thin compatibility layer to isolate internal API changes, providing largely automatic forward support for future versions.

\paragraph{\coq\ Backend.}
For \coq, we build on top of \texttt{coq\_serapy}\footnote{\href{https://github.com/HazardousPeach/coq\_serapy}{https://github.com/HazardousPeach/coq\_serapy}} \cite{sanchez2020generating}, which provides structured access to \coq's proof engine. Our backend extends this infrastructure to support systematic extraction of proof states and tactic traces. The system supports \coq\ (formerly \textsc{Coq}) versions 8.10 through 8.18 and maps assistant-specific state representations into the same unified format used for \lean. Although the internal interaction mechanism differs from the \lean\ meta-programmed backend, both assistants expose a consistent abstract interface to the rest of the framework.

\paragraph{Environment Pooling and Parallel Execution.}
A key feature of \texttt{itp-interface} is the ability to clone proof environments and maintain pools of independent instances initialized to identical states. This abstraction enables concurrent tactic execution and forms the foundation of our parallel proof search module. The pool can be integrated with distributed execution frameworks such as Ray, enabling scalable proof search and large-scale proof annotation in both \lean\; and \coq.

\paragraph{Declaration Database and Tooling.}
Beyond tactic tracing, the interface layer extracts declaration metadata and dependency edges into a structured SQLite database. We provide a Streamlit-based \emph{\lean\ Declaration Database Explorer} for interactive querying, dependency visualization, and connected-component analysis of large formal repositories. This tooling supports repository-level inspection, empirical studies of formal developments, and proof engineering workflows.

\paragraph{Generality and Limitations Across ITPs.}
While the interface abstraction is designed to be assistant-agnostic, our large-scale experiments focus on \lean\; and \coq, which share a tactic-based interaction model. Extending this abstraction to ITPs with fundamentally different interaction paradigms introduces additional challenges. For example, Isabelle follows a document-oriented architecture in which proofs are processed as structured theory files rather than through stepwise tactic execution. 

We implemented preliminary support for Isabelle via the PISA server, which exposes a JSON-RPC interface compatible with our framework. However, in practice, this integration incurs substantial system-level overhead: each PISA instance requires a full Isabelle installation and heap image (tens of gigabytes), leading to prohibitive memory and storage requirements when scaling to many parallel environments. In contrast to the lightweight environment cloning used for \lean\; and \coq, this makes large-scale parallel proof search impractical with current tooling.

Despite these limitations, our core abstractions—standardized state representations, tactic traces, and data formats—remain compatible with such systems. We therefore view full support for document-oriented assistants like Isabelle as an engineering challenge rather than a conceptual limitation, and leave scalable integration as an important direction for future work.

\subsection{Parallel Proof Search Module}
\label{sec:searching-module}

The proof search module operates over the state-transition system defined in \Cref{sec:problem-formulation}. Given a proof state $O$, the search procedure generates candidate tactics and applies them to produce successor states. We implement beam search and best-first search strategies over this state space.

\paragraph{Environment Pooling and Parallel Execution.}
A central contribution of this module is its integration with the environment pool abstraction introduced in \Cref{sec:interface-module}. Instead of executing candidate tactics sequentially within a single ITP instance, we dispatch tactic applications across multiple cloned proof environments. Each environment instance maintains an independent proof state. The pool dynamically filters instances to match frontier states during search and creates additional instances when needed.

This design enables concurrent exploration of multiple branches of the proof tree. When integrated with Ray~\cite{moritz2018ray}, the pool can distribute tactic execution across processes or machines, allowing scalable proof search in both \lean\; and \coq.

\paragraph{Parallel Beam Search.}
\Cref{fig:beam-search-code} presents pseudocode for our parallel beam search implementation. The search maintains a frontier of states, generates candidate tactics using a guidance model, and executes these tactics in parallel across matching environment instances. The pool ensures that only environments whose internal state matches the current frontier state are reused, avoiding unnecessary reinitialization overhead.

\newcommand{\Initialize}{\proc{Initialize}}
\newcommand{\GenerateProofSteps}{\proc{GenerateProofSteps}}
\newcommand{\Execute}{\proc{ExecuteParallel}}
\newcommand{\Timeout}{\mathit{t}}
\newcommand{\Model}{\mathit{model}}
\newcommand{\Width}{\mathit{width}}
\newcommand{\ItpInterfacePool}{\proc{ItpInterfacePool}}
\newcommand{\Pool}{\mathit{pool}}
\newcommand{\Frontier}{\mathit{frontier}}
\newcommand{\TimeElapsed}{\proc{TimeElapsed}}
\newcommand{\ProofTree}{\mathit{proof\_tree}}

% \begin{figure}[H]
% \begin{codebox}
% \Procname{$\proc{ParallelBeamSearch}$($O_0$, $\Model$, $\Timeout$, $\Width$)}
% \li $\Pool \gets \ItpInterfacePool.\Initialize(O_{0})$
% \li $\Frontier \gets \{O_{0}\}$
% \li $\ProofTree \gets \phi$
% \li \While $\Frontier \ne \phi$
% \li \Do
%     \li \If $\TimeElapsed(\Timeout)$
%         \li \Then \Return $\proc{False}$, $\ProofTree$
%     \li $\mathbb{O} \gets \phi$
%     \li \For $O \in \Frontier$
%         \li $\mathbb{A} \gets \GenerateProofSteps(O, \Model, \Width)$
%         \li $\Pool' \gets \Pool.\proc{Filter}(O)$
%         \li \If $\Pool'$ is \textbf{empty}
%             \li \Then $\Pool' \gets \ItpInterfacePool.\Initialize(O)$
%             \li $\Pool.\proc{Merge}(\Pool')$
%         \End
%         \li $\mathbb{O} \gets \mathbb{O} \land \Pool'.\Execute(\mathbb{A})$
%         \li \If $\QED \in \mathbb{O}$
%             \li \Then \Return $\proc{True}$, $\ProofTree$
%         \End
%     \End
%     \li $\Frontier \gets \mathbb{O}.\proc{TopK}(\Width)$
% \End
% \li \Return $\proc{False}$, $\ProofTree$
% \end{codebox}
% \caption{
% Parallel beam search using an environment pool. Candidate tactics are executed concurrently across cloned ITP instances. The pool filters environments to match frontier states and creates new instances when required.
% }
% \label{fig:beam-search-code}
% \end{figure}

Compared to sequential search frameworks (e.g., \pantograph \cite{pantograph}, LeanDojo \cite{yang2023leandojo} for \lean\;), our design avoids repeatedly executing tactics in a single process and instead explores multiple branches of the proof tree simultaneously. This improves throughput and enables scalable experimentation often needed for more challenging theorems.

\paragraph{Proof Tree Annotation and Visualization.} During search, we maintain an annotated proof tree recording states, tactic applications, and transition scores. Each node corresponds to a proof state, and each edge corresponds to a successfully executed tactic. Because tactics are only added when they execute without error in the underlying ITP, the tree represents valid state transitions in the proof environment.

These annotated trees serve multiple purposes: (1) qualitative analysis of search behavior, (2) visualization of explored proof branches, (3) debugging and proof engineering, and (4) potential training signals for iterative refinement methods.

\begin{figure}[H]
\centering
\begin{subfigure}{1.0\linewidth}
    \centering
    \includegraphics[scale=0.15]{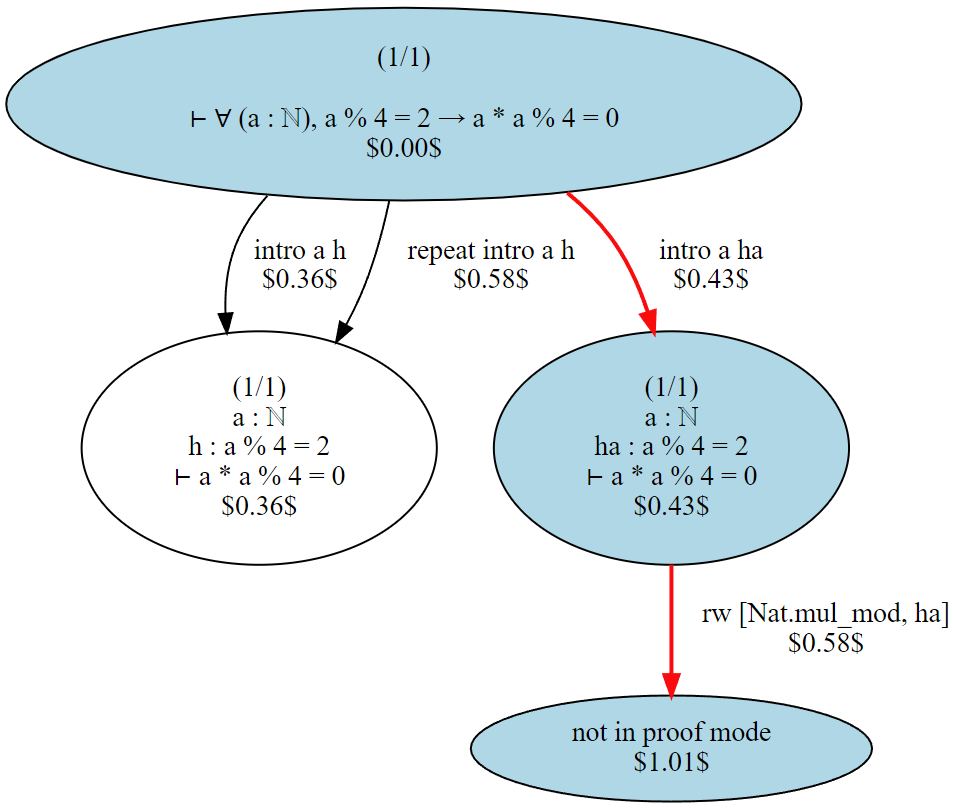}
    \caption{}
\end{subfigure}

\vspace{0.25em}

\begin{subfigure}{1.0\linewidth}
    \centering
    \includegraphics[scale=0.35]{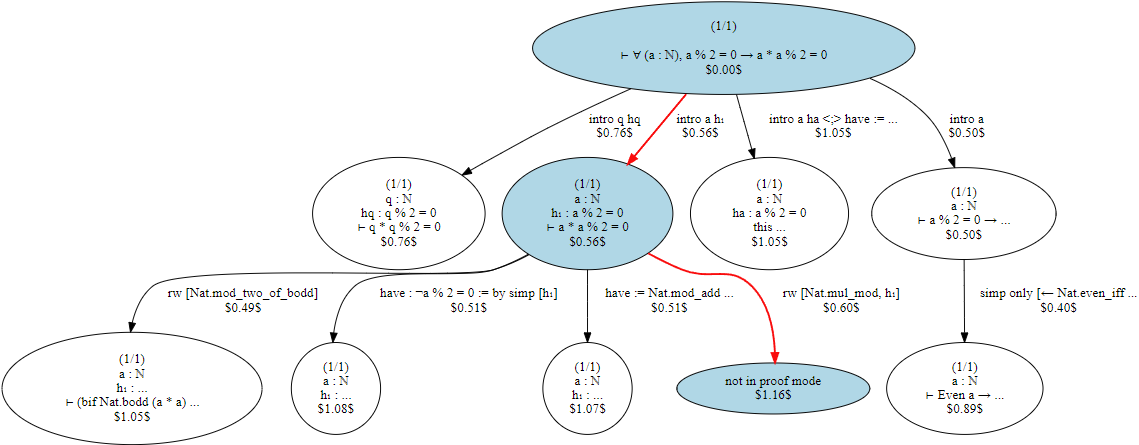}
    \caption{}
\end{subfigure}

\caption{
Visualization of proof trees generated by the parallel proof search module for \lean\; theorems:
(a) $\forall (a : \mathbb{N}), a\;\%\;4 = 2 \rightarrow a * a\; \%\;4 = 0$,
(b) $\forall (a : \mathbb{N}), a\;\%\;2 = 0 \rightarrow a * a\; \%\;2 = 0$.
Nodes represent proof states; edges correspond to successfully executed tactics. Edge labels (within '\$') denote tactic scores.
}
\label{fig:proof-search-annotation}
\end{figure}

\Cref{fig:proof-search-annotation} shows example proof trees generated during beam search for \lean\; theorems. Edge labels indicate tactic scores (e.g., negative log-likelihood under a guidance model), and the correct proof path can be highlighted within the explored tree.

%\wrapfigure{}{}
%\begin{wrapfigure}{r}{3.7in}

\section{Dataset \& Model\\Details}
\newcommand{\CompCertUrl}[0]{\href{https://github.com/AbsInt/CompCert}{https://github.com/AbsInt/CompCert}}
\newcommand{\MathCompUrl}[0]{\href{https://github.com/math-comp/math-comp}{https://github.com/math-comp/math-comp}}
\newcommand{\GeoCoqUrl}[0]{\href{https://github.com/GeoCoq/GeoCoq}{https://github.com/GeoCoq/GeoCoq}}
\newcommand{\CategoryTheoryUrl}[0]{\href{https://github.com/jwiegley/category-theory}{https://github.com/jwiegley/category-theory}}
\newcommand{\LeanUrl}[0]{\href{https://github.com/leanprover-community/mathlib4}{https://github.com/leanprover-community/mathlib4}}
\newcommand{\CatTheory}[0]{\textsc{Cat-Theory}}

\begin{wrapfigure}{r}{3.7in}
\vspace{-1.2in}
\footnotesize
\begin{codebox}
\Procname{$\proc{ParallelBeamSearch}(O_0, \Model, \Timeout, \Width)$}

\li \Comment{Initialize pool with environments at initial state}
\li $\Pool \gets \ItpInterfacePool.\Initialize(O_0)$
\li $\Frontier \gets \{O_0\}$
\li $\ProofTree \gets \varnothing$

\li \While $\Frontier \neq \varnothing$
\li \Do
    \li \If $\TimeElapsed(\Timeout)$
        \li \Then \Return $\proc{False}, \ProofTree$
    \End

    \li $\mathbb{O} \gets \varnothing$ \Comment{Next frontier states}

    \li \For $O \in \Frontier$ 
    \li \Do
        \li $\mathbb{A} \gets \GenerateProofSteps(O, \Model, \Width)$
        
        \li \Comment{Reuse environments already at state $O$}
        \li $\Pool' \gets \Pool.\proc{Filter}(O)$

        \li \If $\Pool'$ is empty
            \li \Then 
                \li $\Pool' \gets \ItpInterfacePool.\Initialize(O)$
                \li \Comment{Register new environments}
                \li $\Pool.\proc{Merge}(\Pool')$ 
        \End

        \li \Comment{Execute candidate tactics in parallel}
        \li $(\mathbb{O}', \Pool') \gets \Pool'.\Execute(\mathbb{A})$

        \li $\Pool.\proc{UpdateState}(\Pool')$ \Comment{Re-index new states}
        \li $\mathbb{O} \gets \mathbb{O} \cup \mathbb{O}'$

        \li \If $\QED \in \mathbb{O}$
            \li \Then \Return $\proc{True}, \ProofTree$
        \End
    \End
    
    \li \Comment{Beam pruning}
    \li $\Frontier \gets \mathbb{O}.\proc{TopK}(\Width)$
    \End
\End

\li \Return $\proc{False}, \ProofTree$
\end{codebox}
\caption{Parallel beam search using an environment pool. Candidate tactics are executed concurrently across cloned ITP instances. The pool maintains a mapping from proof states to environment instances and is updated after each execution step.}
\vspace{-0.7in}
\label{fig:beam-search-code}
\end{wrapfigure}

This section describes the datasets constructed using the \proofwala\ extraction pipeline and the training configurations used to validate the framework. The goal is to document the scale, diversity, and structure of the extracted datasets, as well as the model configurations built on top of them. Experimental results are presented separately in \Cref{sec:experiments}.

\subsection{Dataset Construction}
\label{sec:dataset-details}

Using the \texttt{itp-interface} module, we extract proof-state–tactic pairs from mature repositories in both \lean\; and \coq. The extraction process replays tactic scripts inside the assistant’s kernel and records semantically valid state transitions in the unified representation described in \Cref{sec:problem-formulation}. This ensures that all extracted data corresponds to valid kernel-level proof execution.

We extract data from the following repositories: (1) \textbf{CompCert} \footnote{\CompCertUrl}, (2) \textbf{MathComp} \footnote{\MathCompUrl}, (3) \textbf{GeoCoq} \footnote{\GeoCoqUrl}, (4) \textbf{Mathlib4} \footnote{\LeanUrl}, and (5) \textbf{CategoryTheory} \footnote{\CategoryTheoryUrl}

For CompCert and Mathlib, we reuse established train/test splits from prior work \cite{sanchez2020generating,yang2023leandojo}. For the remaining repositories, we construct theorem-level splits, ensuring at least 500 test theorems when possible.

\begin{table*}[t]
\centering
\small
\setlength{\tabcolsep}{5pt}
\renewcommand{\arraystretch}{1.15}

% ===================== SUBTABLE (a) =====================
\begin{subtable}{\textwidth}
\centering
\scalebox{0.7}{
\begin{tabular}{llll}
\toprule
\multicolumn{4}{c}{\textbf{Initial Fine-tuning}} \\
\midrule
\textbf{Data-mix} & \textbf{Source} & \textbf{Model Trained} & \textbf{Token Count} \\
\midrule
1. CompCert & CompCert Repo & -- & 61.6M \\
2. MathComp & MathComp Repo & -- & 18.2M \\
3. GeoCoq & GeoCoq Repo & -- & 91.2M \\
4. \coq & Mix 1--3 & \coq & 171M \\
5. \lean & Mathlib Repo & \lean & 99M \\
6. \multi & Mix 4--5 & \multi & 270M \\
\midrule
\multicolumn{4}{c}{\textbf{Further Fine-tuning}} \\
\midrule
7. CategoryTheory & CategoryTheory Repo & 
\multi-\CatTheory\; \&\; \coq-\CatTheory & 1.7M \\
\bottomrule
\end{tabular}}
\caption{Data-mixes constructed using the unified \proofwala\ extraction pipeline.}
\label{tab:data-mix}
\end{subtable}

\vspace{1em}

% ===================== SUBTABLE (b) =====================
\begin{subtable}{\textwidth}
\centering
\scalebox{0.7}{
\begin{tabular}{lrrrrrr}
\toprule
& \multicolumn{3}{c}{\textbf{\# Proof-Step \& State Pairs}} 
& \multicolumn{3}{c}{\textbf{Theorem Count}} \\
\cmidrule(lr){2-4}\cmidrule(lr){5-7}
\textbf{Data-mix} 
& \textbf{Train} & \textbf{Test} & \textbf{Val}
& \textbf{Train} & \textbf{Test} & \textbf{Val} \\
\midrule
1. CompCert & 80288 & 6199 & -- & 5440 & 501 & -- \\
2. MathComp & 34196 & 1378 & 2285 & 11381 & 536 & 729 \\
3. GeoCoq & 91120 & 12495 & 4928 & 4036 & 505 & 208 \\
4. \coq & 205604 & 20072 & 7213 & 20857 & 1542 & 937 \\
5. \lean & 237003 & 4323 & 4220 & 56140 & 991 & 1035 \\
6. \multi & 442607 & 24395 & 11433 & 76997 & 2533 & 1972 \\
7. CategoryTheory & 4114 & 610 & 208 & 573 & 101 & 43 \\
\bottomrule
\end{tabular}}
\caption{Dataset sizes under the unified \proofwala\ representation.}
\label{tab:data-mix-size}
\end{subtable}

\caption{Overview of multilingual datasets constructed and used for training and evaluation.}
\label{tab:dataset-overview}
\vspace{-0.1in}
\end{table*}

% \renewcommand\theadfont{}
% \begin{wraptable}{r}{0.6\textwidth}
%     %\vspace{-0.1in}
%     \centering
%     \scalebox{0.55}{
%     \begin{tabular}{llll}
%     \toprule
%     \multicolumn{4}{c}{\thead{\textbf{Initial Fine-tuning}}}\\
%     \hline
%     \thead{\textbf{Data-mix}}  & \thead{\textbf{Data-mix Source}}  & \thead{\textbf{\name\;}\\\textbf{Models Trained}} & \thead{\textbf{Token Count}}\\
%     \toprule
%     1. CompCert & CompCert Repo & - & 61.6 M \\
%     2. MathComp & MathComp Repo & - & 18.2 M\\
%     3. GeoCoq & GeoCoq Repo & - & 91.2 M\\
%     4. \coq & {\textbf{Data-Mixes:} 1-3} & \coq & 171 M \\
%     5. \lean & Mathlib Repo & \lean & 99 M \\
%     6. \multi & {\textbf{Data-Mixes:} 4-5} & \multi & 270 M \\
%     \hline
%     \multicolumn{4}{c}{\thead{\textbf{Further Fine-tuning}}}\\
%     \hline
%     7. CategoryTheory & CategoryTheory Repo & \thead{\multi-\CatTheory\;\\ \&\; \coq-\CatTheory} & 1.7 M\\
%     \bottomrule
%     \end{tabular}}
%     \caption{Data-mixes extracted using the unified \proofwala\ format.}
%     \label{tab:data-mix}
%     %\vspace{-0.5in}
% \end{wraptable}

Across all repositories, we extract \texttt{442,607} proof-step/state pairs from \texttt{76,997} theorems spanning both \lean\; and \coq. The CategoryTheory dataset is excluded from initial training and reserved for domain adaptation experiments.

To evaluate different training configurations, we construct several data-mixes combining monolingual and multilingual datasets. These are summarized in \Cref{tab:data-mix}.

% \begin{wraptable}{r}{0.6\textwidth}
%     %\vspace{-0.4in}
%     \centering
%     \scalebox{0.5}{
%     \begin{tabular}{lrrrrrr}
%     \hline
%     & \multicolumn{3}{c}{\textbf{\# Proof-Step \& State Pairs}}  & \multicolumn{3}{c}{\textbf{Theorem Count}}\\
%     \cmidrule(lr){2-4}\cmidrule(lr){5-7}
%     \textbf{Data-mix} & \textbf{Train} & \textbf{Test} & \textbf{Val} & \textbf{Train} & \textbf{Test} & \textbf{Val}\\
%     \toprule
%     1. CompCert & 80288 & 6199 & - & 5440 & 501 & - \\
%     2. MathComp & 34196 & 1378 & 2285 & 11381 & 536 & 729 \\
%     3. GeoCoq & 91120 & 12495 & 4928 & 4036 & 505 & 208 \\
%     4. \coq & 205604 & 20072 & 7213 & 20857 & 1542 & 937 \\
%     5. \lean & 237003 & 4323 & 4220 & 56140 & 991 & 1035 \\
%     6. \multi & 442607 & 24395 & 11433 & 76997 & 2533 & 1972 \\
%     7. CategoryTheory & 4114 & 610 & 208 & 573 & 101 & 43 \\
%     \bottomrule
%     \end{tabular}}
%     \caption{Size of datasets extracted using the unified \proofwala\ representation.}
%     \label{tab:data-mix-size}
%     \vspace{-0.8in}
% \end{wraptable}

\Cref{tab:data-mix-size} reports the number of extracted proof-step/state pairs and theorem counts for each data-mix. These statistics illustrate the scale and diversity of the extracted multilingual dataset.

\subsection{Training Data Representation for Proof Step Prediction}
\label{sec:training-data}

The extracted proof traces are stored in a unified \texttt{JSON} format that abstracts over assistant-specific details while preserving full semantic information about proof states and tactic transitions. Each training example corresponds to a transition $(O_i, a_i, O_{i+1})$ as defined in \Cref{sec:problem-formulation}, where: (1) $O_i$ (start\_goals) represents the current set of proof obligations, (2) $a_i$ (proof\_steps) represents the applied tactic, and (3) $O_{i+1}$ (end\_goals) represents the resulting obligations.

An excerpt of the extracted data format for \coq\; and \lean\; 4 is shown in \Cref{fig:training-data-collection-format}. Although the underlying assistants differ in syntax and internal representations, the stored format is structurally identical across languages.

\begin{figure}[H]
\begin{mdframed}[roundcorner=5pt]
\begin{minipage}{0.45\linewidth}
(a)
\begin{minted}[breaklines]{json}
{
"theorem_name": "nat_plus_0_is_n"
"start_goals": [
    {
        "hypotheses": [],
        "goal": "forall n : nat, n + 0 = n",
        "# ... extra metadata"
    }
],
"proof_steps": ["intros n."],
"end_goals": [
    {
        "hypotheses": ['n: nat'],
        "goal": "n + 0 = n",
        "# ... extra metadata"
    }
],
"# ... extra metadata"
} 
\end{minted}
\end{minipage}
\begin{minipage}{0.45\linewidth}
(b)
\begin{minted}[breaklines]{json}
{
"theorem_name": "nat_plus_0_is_n"
"start_goals": [
    {
        "hypotheses": [],
        "goal": "∀ (n : Nat), Nat.add n 0 = n",
        "# ... extra metadata"
    }
],
"proof_steps": ["intro n"],
"end_goals": [
    {
        "hypotheses": ['n : Nat'],
        "goal": "Nat.add n 0 = n",
        "# ... extra metadata"
    }
],
"# ... extra metadata"
} 
\end{minted}
\end{minipage}
\end{mdframed}
\caption{Excerpt from the extracted proof-step data in unified JSON format. Each entry represents a state transition $(O_i, a_i, O_{i+1})$. The format is identical across \coq\; and \lean\; 4. Additional metadata fields (not shown) include file paths, namespaces, and dependency information.}
\label{fig:training-data-collection-format}
\end{figure}

From this unified JSON representation, we construct standardized prompts for proof step prediction. The prompt encodes the current proof state (goals and hypotheses) and requests the next tactic. Importantly, the prompt format is identical across \lean\; and \coq, and does not reveal the originating assistant or repository.

This design enables cross-assistant training and evaluation without introducing assistant-specific tokens into the model’s input.

\subsection{Model Configuration}
\label{sec:model-details}

To validate the usability of the extracted datasets, we fine-tune a pretrained \codeTFive-\base\ model (220M parameters) for proof step prediction.

We train three configurations: (1) \proofwala-\coq\ (trained on \coq\ repositories), (2) \proofwala-\lean\ (trained on \lean\ repositories), and (3) \proofwala-\multi\ (trained jointly on both assistants).

All models are trained for the same number of optimization steps and batch sizes to ensure comparability across configurations. Hyperparameters are provided in \Cref{tab:hyperparams}.

These trained models are used in the parallel proof search module described in \Cref{sec:searching-module}. Experimental results evaluating these models are presented in \Cref{sec:experiments}.

\begin{wraptable}{r}{3in}
    \centering
    % \vspace{-0.8in}
    \scalebox{0.7}{
    \begin{tabular}{ll}
    \hline
    \thead{\textbf{Hyperparameter}}  & \thead{\textbf{Value}}\\
    \toprule
    \textbf{Pretrained Model Name} & \codeTFive-\base\;(220 M) \\
    \textbf{Learning Rate} & $2 \times 10^{-4}$\\
    \textbf{Learning Scheduler Type} & \texttt{cosine}\\
    \textbf{Warmup Ratio} & \texttt{0.03}\\
    \textbf{Weight Decay} & \texttt{0.001}\\
    \textbf{Max Grad Norm (Gradient Clipping)} & \texttt{0.3}\\
    \textbf{Optimizer} & \texttt{adamw\_torch}\\
    \textbf{Gradient Accumulation Steps} & \texttt{1}\\
    \textbf{Max \# Steps (Gradient Updates)} & 34000\\
    \textbf{Batch Size} & 128\\
    \textbf{Checkpoint \# Steps} & 20000\\
    \textbf{Max \# Tokens} & 2048\\
    \bottomrule
    \end{tabular}}
    \caption{\footnotesize{Hyperparameters used for training our \proofwala-\{\lean, \coq, \multi\}. In line with recent work on training multilingual models for autoformalization \cite{jiang2023multilingualmathematicalautoformalization}, we used the same step count and batch sizes to train all our models on different data mixes ensuring our ablation studies about the transfer were fair and were not merely a result of training more on bigger data-mixes.}}
    \vspace{-0.25in}
    \label{tab:hyperparams}
\end{wraptable}

\paragraph{Model choice and scope of evaluation.}
All experiments in this work use \codeTFive-\base\ (220M parameters) as the underlying proof-step prediction model. This choice is driven by our experimental goals. First, we train multiple models across different data-mixes (monolingual and multilingual) to study transfer effects under controlled conditions; using a moderate-sized model allows us to keep training budgets comparable across these settings. Second, our primary objective is to evaluate the capabilities of the \proofwala\ framework---including unified data extraction, training, and parallel proof search---rather than to optimize absolute theorem-proving performance. Since the framework is model-agnostic and built on top of standard \textsc{HuggingFace} training pipelines, its functionality and scalability are largely independent of the specific model size.

While larger models and more complex systems (e.g., with retrieval or verifier augmentation) achieve stronger results on benchmark datasets, we intentionally use a fixed-capacity model to isolate the effect of multilingual training on proof-step prediction and search. The improvements observed for the multilingual model over monolingual baselines therefore reflect data-driven transfer rather than gains from increased model capacity. We expect these transfer effects to persist---and potentially strengthen---when scaling to larger models, although verifying this empirically remains an important direction for future work.

\section{Evaluation}
\newcommand{\cmark}{\ding{51}}%
\newcommand{\xmark}{\ding{55}}%
\newcommand{\codetfive}[0]{\textsc{CodeT5}}
\newcommand{\greencheck}{\color{green}{\checkmark}}
\newcommand{\redcross}{\color{red}{\xmark}}

Using our trained \proofwala{} models, we evaluate (i) the benefit of incorporating multilingual proof data into proof-step prediction and end-to-end proof search, and (ii) whether multilingual pretraining improves adaptation to a novel domain after further fine-tuning. 

Because \proofwala\ unifies assistant-level execution, proof-state extraction, and annotated search-tree logging across \lean\; and \coq, it enables not only aggregate evaluation (pass-at-$k$), but also fine-grained analysis of proof search dynamics. All experiments use the \proofwala-\{\multi, \coq, \lean\} models inside the search module described in \Cref{sec:searching-module}. We run proof search on the test splits defined in \Cref{tab:data-mix-size} for CompCert, MathComp, GeoCoq, CategoryTheory, and \lean.

\subsection{Experimental Setup}
\label{sec:experiments}

\paragraph{Search procedure.}
We use \proofwala\ proof-step models for single-step tactic prediction and conduct end-to-end proof search via beam search. Candidate tactics are ranked using the negative log-likelihood assigned by the proof-step model, and this score is used as the primary heuristic for node expansion. We employ a timeout of \texttt{600} seconds for most experiments; for GeoCoq, we use a higher timeout of \texttt{1200} seconds due to longer proof scripts and higher tactic execution latency.

\paragraph{Parallel execution.}
All proof search experiments use the environment-pool abstraction of \texttt{itp-interface}. 
Each frontier state is mapped to one or more cloned assistant environments, allowing candidate tactics to be executed in parallel without reinitializing the proof context. 
Because environment instances retain assistant-level proof states, this design preserves semantic correctness while enabling scalable exploration. 
 The abstraction is shared across \lean\; and \coq\ backends (see \Cref{sec:interface-module} and \Cref{sec:searching-module}). 
 
%  Search hyperparameters are summarized in \Cref{tab:search-params}.
% \todo{should we add search params??}

\paragraph{Ablations and statistical testing.}
To study transfer, we compare monolingual models (\proofwala-\coq, \proofwala-\lean) against the multilingual model (\proofwala-\multi) on each data-mix. We additionally report paired bootstrap significance testing for the gap between multilingual and monolingual variants where applicable.

\subsection{Aggregate Results}
\label{sec:aggregate-results}

\Cref{tab:all-experiments} summarizes \emph{pass-at-$k$} results for $1 \le k \le 5$ across all data-mixes. Overall, \proofwala-\multi\ matches or improves over monolingual baselines across both assistants, suggesting that multilingual training provides useful transfer signals for proof-step prediction and downstream proof search. We observe statistically significant improvements on the largest benchmark (\lean/\textsc{Mathlib}), while improvements in other datasets are smaller and not statistically significant, and should be interpreted as indicative trends.

To evaluate adaptation to a new domain, we further fine-tune \proofwala-\multi\ and \proofwala-\coq\ on CategoryTheory. The multilingual model achieves substantially higher pass-at-$k$ after fine-tuning, with statistically significant improvements over the \coq-only baseline, indicating that multilingual pretraining can improve downstream adaptation even when the target domain is represented in only one assistant. 

This behavior is consistent with prior work on multilingual mathematical reasoning. In particular, MMA~\cite{jiang2023multilingualmathematicalautoformalization} demonstrates that models trained on multiple formal languages achieve improved downstream performance compared to monolingual training. Our results extend this observation to the setting of proof-step prediction and neural-guided proof search.

\begin{table}
    \centering
    \scalebox{0.65}{
    \begin{tabular}{lclllllll}
    \toprule
    \multicolumn{2}{c}{\textbf{Data-Mix}} &  
     & 
    \multicolumn{5}{c}{\textbf{Pass-at-$k$} \%} &
    \\
    \cmidrule(lr){1-2}\cmidrule(lr){4-8}
    \textbf{Name} & 
    \textbf{\# Theorems} & 
    \thead[c]{\textbf{Proof Step Model}} &
    \textbf{Pass@1} & 
    \textbf{Pass@2} & 
    \textbf{Pass@3} & 
    \textbf{Pass@4} & 
    \textbf{Pass@5} & 
    \thead[c]{\textbf{$p_{\mathrm{value}}$}\\(\textbf{$\alpha$}: 0.05)\footref{fnote:p-value}}\\
    \toprule
    \textbf{\lean} & 
    991 &
    \proofwala-\lean & 
    24.92 & 
    26.64 & 
    27.54 &
    28.05 &
    28.25 &
    \\
     & 
     & 
     \proofwala-\multi & 
     \textbf{26.84} & 
     \textbf{28.56} & 
     \textbf{29.67} &
     \textbf{29.97} &
     \textbf{30.58} &
     \textbf{0.018} \\
    \hline
    \textbf{MathComp} & 
    536 & 
    \proofwala-\coq & 
    \textbf{28.28} & 
    28.65 & 
    29.4 &
    29.59 &
    30.15 &
    \\
     &
     &
     \proofwala-\multi & 
     27.9 & 
     \textbf{29.21} & 
     \textbf{29.59} &
     \textbf{30.15} &
     \textbf{30.52} &
     0.355
     \\
    \hline
    \textbf{GeoCoq} & 
    505 & 
    \proofwala-\coq & 
    \textbf{32.87} & 
    \textbf{33.66} & 
    33.86 &
    34.06 &
    34.46 &
    \\
    & 
    & 
    \proofwala-\multi & 
    30.89 & 
    \textbf{33.66} & 
    \textbf{34.65} &
    \textbf{35.64} &
    \textbf{35.84} &
    0.135
    \\
    \hline
    \textbf{CompCert} &
    501 & 
    \proofwala-\coq &  
     17.56 & 
     18.76 & 
     19.16 &
     19.76 &
     20.76 &
     \\
     &
     & 
     \proofwala-\multi & 
     \textbf{17.96} & 
     \textbf{19.76} & 
     \textbf{20.56} &
     \textbf{21.16} &
     \textbf{21.96} &
     0.191 \\
    \hline
    \textbf{CategoryTheory} & 
    101 & 
    \proofwala-\coq-\CatTheory & 
    36.63 & 
    42.57 & 
    44.55 & 
    44.55 & 
    45.54 & \\
    & 
    & 
    \proofwala-\multi-\CatTheory & 
    \textbf{44.55} & 
    \textbf{51.49} & 
    \textbf{52.48} & 
    \textbf{53.47} & 
    \textbf{53.47} & 
    \textbf{0.008} \\
    \bottomrule
    \end{tabular}}
    \caption{Pass-at-$k$ results across data-mixes using \proofwala\ proof search guided by different proof-step models. \proofwala-\multi\ is trained jointly on \lean\; and \coq\ data; monolingual baselines are trained only on their respective assistants. CategoryTheory results report further fine-tuning on that domain.}
    \label{tab:all-experiments}
    \vspace{-0.2in}
\end{table}
\addtocounter{footnote}{+1}
\footnotetext{\label{fnote:p-value}Statistical significance is assessed using a paired bootstrap test; we report significance when $p_{\mathrm{value}} < 0.05$.}

\subsection{Case Study: Cross-Lingual Transfer in Category Theory}
\label{sec:cat-theory-transfer}

To better understand the transfer gains in CategoryTheory, we examined \coq{} theorems that the multilingual model successfully proved while the \coq{}-only model failed. In many such cases, we found structurally analogous \lean\; theorems in the multilingual training data, particularly for categorical constructions such as adjunctions, monicity, currying, and evaluation morphisms. Because the framework extracts theorem metadata and proof traces across assistants in a unified format, we can systematically identify structurally analogous theorems between \coq\ and \lean\ and analyze transfer behavior at the declaration level. \Cref{tab:cat-theory-transfer} shows representative examples.

This result is especially interesting in light of prior findings in multilingual NLP. For instance, XLM-R \cite{conneau2019unsupervised} shows that multilingual training can lead to negative transfer, particularly for low-resource languages, where out-of-language tokens may dilute the model’s effectiveness. By contrast, we observe that even when training on mixed-language data (with $M > N$ tokens), our multilingual models consistently outperform monolingual counterparts trained on the smaller in-language subset ($N$ tokens). This suggests that formal languages exhibit more meaningful structural alignment across ITPs, enabling productive cross-lingual transfer rather than suffering from the interference effects observed in natural language settings.

\begin{table}[t]
\centering
\scalebox{0.85}{
\renewcommand{\arraystretch}{1.2}
\begin{tabular}{|p{4.2cm}|p{7.0cm}|c|c|}
\hline
\textbf{\coq\ Theorem} & \textbf{\lean\ Equivalent} & \textbf{Multilingual} & \textbf{\coq-Only} \\
\hline
\texttt{counit\_fmap\_unit} \newline $\forall\; x,\; \varepsilon \circ \mathsf{fmap}[F]\; \eta \approx \mathsf{id}[F(x)]$
& \texttt{adjointify\_$\eta$\_$\varepsilon$ (X : C)} \newline 
$F.\mathsf{map}((\mathsf{adjointify}_{\eta, \varepsilon}).\mathsf{hom.app}(X)) \gg \varepsilon.\mathsf{hom.app}(F.\mathsf{obj}(X)) = \mathbb{1}(F.\mathsf{obj}(X))$
& \cmark & \xmark \\
\hline
\texttt{id\_monic} \newline $\forall\; x,\; \mathsf{Monic}(\mathsf{id}_{x})$
& \texttt{cancel\_mono\_id (f : X -> Y)} \newline 
$g \gg f = f \;\Leftrightarrow\; g = \mathbb{1}_X$
& \cmark & \xmark \\
\hline
\texttt{eval\_first} \newline $\mathsf{eval} \circ \mathsf{first}(f) \approx \mathsf{uncurry}(f)$
& \texttt{uncurry\_id\_eq\_ev (A X : C)} \newline $\mathsf{uncurry}(\mathbb{1}_{A \Rightarrow X}) = (\mathsf{exp.ev}(A)).\mathsf{app}(X)$
& \cmark & \xmark \\
\hline
\end{tabular}}
\caption{Representative CategoryTheory examples where the multilingual model succeeds but the \coq-only model fails, alongside structurally analogous \lean\; theorems present in the multilingual training data.}
\label{tab:cat-theory-transfer}
\vspace{-0.25in}
\end{table}

\subsection{Proof Tree Analysis}
\label{sec:proof-tree-analysis}

Beyond pass-at-$k$, \proofwala\ exposes the internal search process via annotated proof trees (\Cref{fig:proof-search-annotation}). Each proof tree contains only compilable transitions, ensuring that every node corresponds to a valid proof state and every edge corresponds to a tactic that executes without error. Because proof states are obtained directly from the assistant after tactic execution, these trees reflect true semantic state transitions rather than simulated or approximate transitions.

\begin{wraptable}{r}{3.2in}
%\begin{table*}
\vspace{-0.3in}
    \centering
    \scalebox{0.6}{
    \begin{tabular}{lllllllll}
    \toprule
    \textbf{Data-Mix} &  
    \textbf{\name\ Model} &
    \multicolumn{3}{c}{\textbf{Avg.\ Proof Tree Stats}}\\
    \cmidrule(lr){3-5}
     & & 
    \textbf{\# Nodes} &
    \textbf{\# Edges} & 
    \textbf{Avg.\ Degree} \\
    \toprule
    \textbf{\lean} & 
    \lean & 
     {3.989} & 
     {3.536} & 
     {1.536}
    \\
     & 
     \multi & 
    4.729 & 
    4.689 & 
    1.983
    \\
    \hline
    \textbf{MathComp} & 
    \coq & 
    2.534 & 
    1.739 & 
    1.167
    \\
     &
     \multi & 
     2.576 & 
     1.822 & 
     1.207 
     \\
    \hline
    \textbf{GeoCoq} & 
    \coq & 
     15.358 & 
     14.457 & 
     1.180
    \\
    & 
    \multi & 
     17.144 & 
     15.75 & 
     1.353
    \\
    \hline
    \textbf{CompCert} &
    \coq &  
     8.048 & 
     7.480 & 
     1.404 
     \\
     &
     \multi & 
     8.318 & 
     8.200 & 
     1.584 
     \\
    \hline
    \textbf{CategoryTheory} & 
    \coq-\CatTheory & 
    5.674 & 
    5.804 & 
    2.301 
    \\
    & 
    \multi-\CatTheory & 
    7.056 & 
    7.130 & 
    2.193
    \\
    \bottomrule
    \end{tabular}}
    \caption{Average proof tree statistics across data-mixes. Proof trees contain only compilable edges and valid states.}
    \vspace{-1.3in}
    \label{tab:tree-stats}
\end{wraptable}
%\end{table*}

\paragraph{Tree size and branching.}
\Cref{tab:tree-stats} compares average proof tree statistics across data-mixes. 
Because the framework logs every successfully executed tactic and resulting proof state, we can quantify structural properties of search, including node count, edge count, and average branching factor. 
In most settings, trees produced by \multi\ are larger and include higher branching factors, suggesting that multilingual models propose a larger number of valid tactics per state.

\paragraph{Search time and proof length.}

\begin{wraptable}{r}{3.5in}
\centering
\vspace{-0.4in}
\scalebox{0.6}{
\small
\setlength{\tabcolsep}{6pt}
\renewcommand{\arraystretch}{1.15}
\begin{tabular}{llrr}
\toprule
\textbf{Data Mix} & \textbf{\name\ Model} & 
\textbf{Avg.\ Proof Length} & 
\textbf{Avg.\ Time (s)} \\
\midrule
\lean      & \multi & 2.0363 & 30.5296 \\
           & \lean  & 2.0679 & 20.7370 \\
\midrule
CompCert   & \multi & 4.7913 & 59.9424 \\
           & \coq   & 5.0270 & 70.3282 \\
\midrule
MathComp   & \multi & 2.4940 & 8.9993  \\
           & \coq   & 2.4759 & 7.5634  \\
\midrule
GeoCoq     & \multi & 9.2486 & 107.1999 \\
           & \coq   & 10.6954 & 74.3914 \\
\midrule
CategoryTheory & \multi & 3.4909 & 39.2182 \\
               & \coq   & 3.0426 & 27.4105 \\
\bottomrule
\end{tabular}}
\caption{
Average proof length (number of tactics) and average proof search time (seconds) across data-mixes. 
Proof lengths are comparable across models, while \multi\ often searches longer, consistent with broader exploration and larger proof trees.
}
\label{tab:proof-length-time-summary}
\vspace{-0.3in}
\end{wraptable}

\Cref{tab:proof-length-time-summary} summarizes both proof length and search time across data-mixes. 
Proof lengths are broadly comparable across models, suggesting that performance differences are not driven by systematically longer proofs. 
However, \multi\ typically searches longer, which correlates with larger proof trees and broader exploration.

The ability to correlate tree size, branching factor, timeout usage, and final proof length is made possible by the framework’s explicit tree representation and execution logging, which records timestamps, state transitions, and tactic outcomes for every search run.

\subsection{Additional Benchmarking and Scalability}
\label{sec:minif2f-scalability}

\paragraph{MiniF2F.}
We additionally evaluate \proofwala\ on the \texttt{MiniF2F} benchmark \cite{zheng2021minif2f,yang2023leandojo}. We run the same search pipeline as in \Cref{sec:experiments}, using \proofwala-\multi{} and \proofwala-\lean{} for \lean\; proof search. On this benchmark, \proofwala-\multi{} achieves \texttt{pass@5} of 26.23\%, slightly exceeding \proofwala-\lean{} (25.41\%), indicating that multilingual training provides consistent gains even when evaluated on an out-of-distribution benchmark.

We emphasize that our goal is not to achieve state-of-the-art performance on MiniF2F. Instead, we use this benchmark as an \emph{out-of-distribution (OOD)} evaluation to study generalization and transfer under a controlled setup. As a result, our absolute performance is lower than recent systems that rely on significantly larger models and additional infrastructure, but the relative improvement of the multilingual model over the monolingual baseline provides evidence of transfer under OOD conditions.

\paragraph{Scalability via parallel environment pools.}
To quantify the impact of parallel proof search, we vary the number of CPU workers used by our Ray-backed environment pool (see \Cref{sec:searching-module}). Increasing parallelism from 8 to 20 CPUs improves MiniF2F \texttt{pass@5} from 22.54\% to 26.23\%, while reducing average proving time from 83.32s to 74.56s. These gains arise from evaluating more candidate tactics per frontier state and pruning unpromising branches earlier. 

Such scaling is enabled by the environment-pool abstraction, which allows assistant instances to be cloned, filtered by proof state, and reused during search rather than restarted from scratch. This architectural design is central to \proofwala: it enables controlled, scalable experimentation on how neural guidance interacts with parallel search, independent of model scale.

% \subsection{Additional Benchmarking and Scalability}
% \label{sec:minif2f-scalability}

% \paragraph{MiniF2F.}
% We additionally evaluate \proofwala\ on the \texttt{MiniF2F} benchmark \cite{zheng2021minif2f,yang2023leandojo}. We run the same search pipeline as in \Cref{sec:experiments}, using \proofwala-\multi{} and \proofwala-\lean{} for \lean\; proof search. On this benchmark, \proofwala-\multi{} achieves \texttt{pass@5} of 26.23\%, slightly exceeding \proofwala-\lean{} (25.41\%), reinforcing that multilingual training can benefit \lean\; proof search even when evaluation is conducted in a single assistant.

% \paragraph{Scalability via parallel environment pools.}
% To quantify the impact of parallel proof search, we vary the number of CPU workers used by our Ray-backed environment pool (see \Cref{sec:searching-module}). Increasing parallelism from 8 to 20 CPUs improves MiniF2F \texttt{pass@5} from 22.54\% to 26.23\%, while reducing average proving time from 83.32s to 74.56s. These gains arise from evaluating more candidate tactics per frontier state and pruning unpromising branches earlier. 
% Such scaling is enabled by the environment-pool abstraction, which allows assistant instances to be cloned, filtered by proof state, and reused during search rather than restarted from scratch. 
% This architectural choice allows us to study how neural guidance interacts with parallel search at scale.

\subsection{Declaration-Level Analysis and Visualization}
\label{sec:declaration-explorer}

Beyond proof search, our framework exposes declaration-level metadata extracted from \lean\; projects, including full dependency graphs, namespace information, file paths, and declaration types. To facilitate interactive analysis, we built a Streamlit-based \emph{Lean Declaration Database Explorer} on top of the extracted SQLite database produced by \texttt{itp-interface}.

\paragraph{Custom SQL-based analysis.}
The explorer supports direct SQL queries over the declaration graph. For example, we can identify declarations with the highest number of dependencies or dependents, enabling structural analysis of core library components. This level of introspection is possible because our Lean integration extracts repository-wide declaration information—not only tactic-level proof traces.

\begin{figure}
    \centering
    \includegraphics[width=\linewidth]{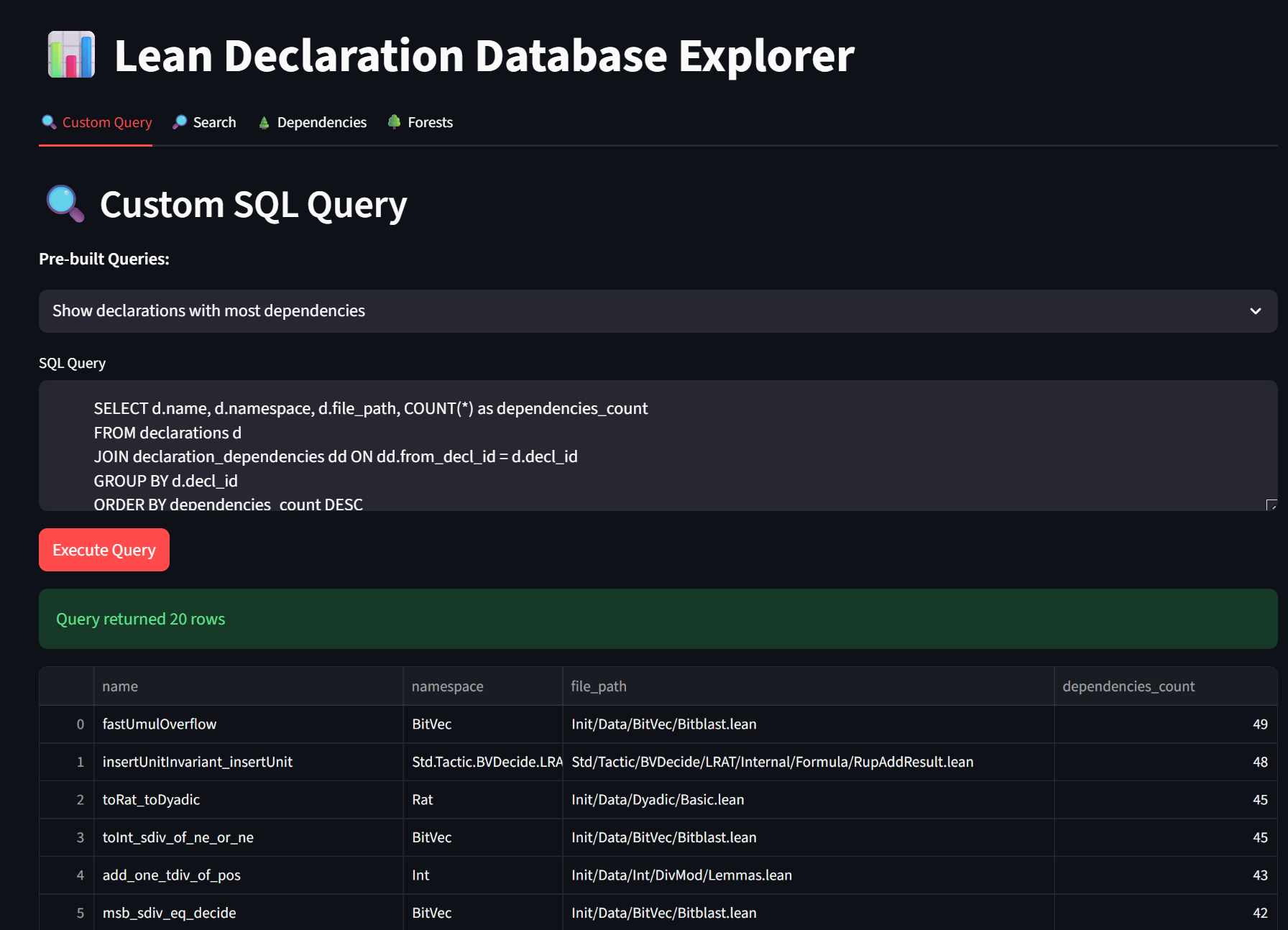}
    \vspace{0.5em}
\includegraphics[width=\linewidth, trim=0 0 40 0, clip]{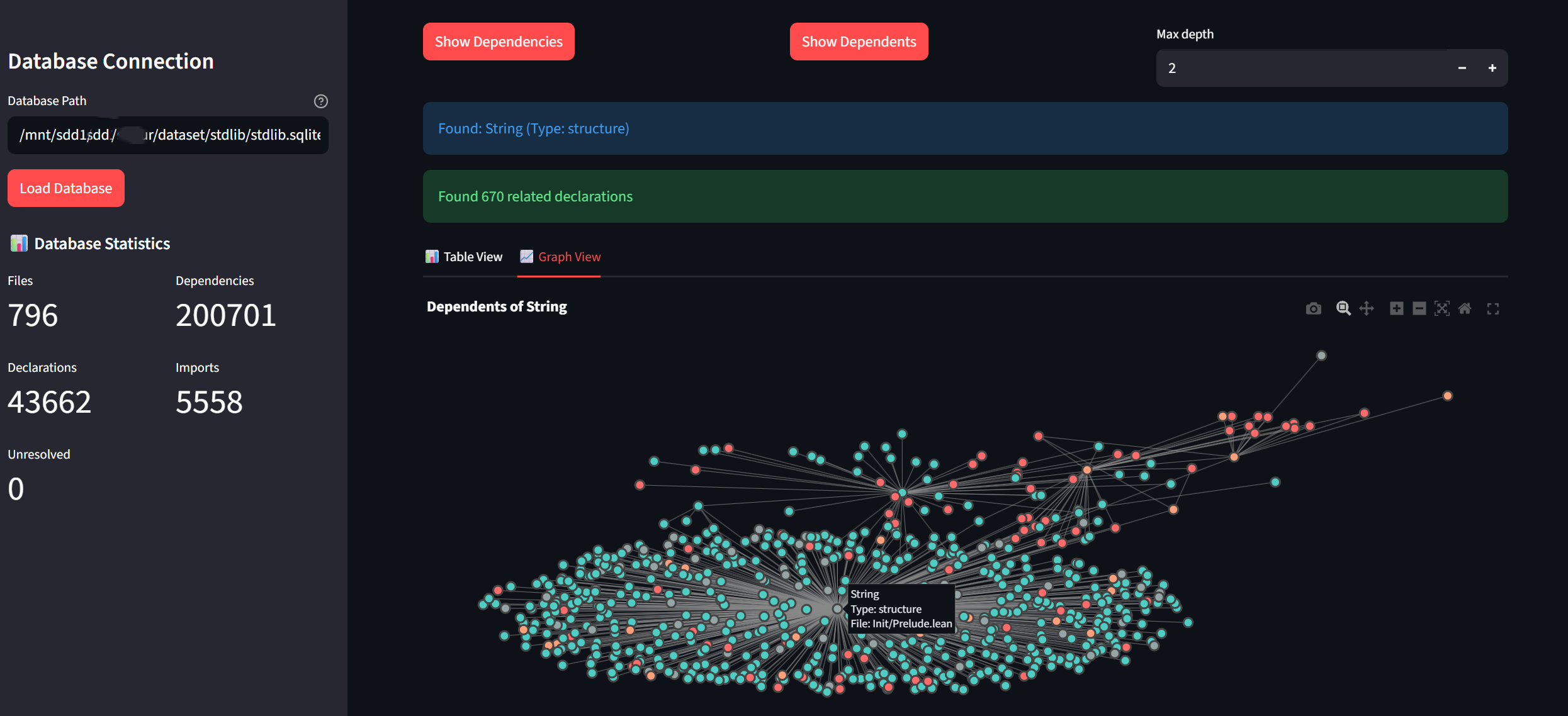}
    \caption{
    Streamlit-based Lean Declaration Database Explorer built on top of \proofwala's declaration extraction pipeline.
    \textbf{Top:} Custom SQL queries over the declaration database, for example identifying declarations with the largest number of dependencies.
    \textbf{Bottom:} Interactive dependency graph visualization for the \texttt{String} type, showing dependent declarations up to a configurable depth.
    Each node corresponds to a declaration, and additional information (name, type, file location) can be inspected interactively by hovering over nodes in the graph.
    }
    \label{fig:streamlit-explorer}
\end{figure}

\paragraph{Dependency graph visualization.}
The explorer also supports interactive visualization of dependency and dependent graphs up to configurable depth. \Cref{fig:streamlit-explorer} shows examples of both custom-query analysis and graph-based dependency exploration.

% \begin{figure}
%     \centering
%     \includegraphics[scale=0.5]{img-streamlit-dep-custom-query.png}
%     \vspace{0.5em}
%     \includegraphics[scale=0.45]{img-streamlit-dep-explorer-final.png}
%     \caption{
%     Streamlit-based Lean Declaration Database Explorer built on top of \proofwala's declaration extraction pipeline. 
%     \textbf{Top:} Custom SQL queries over the declaration database, e.g., identifying declarations with the most dependencies. 
%     \textbf{Bottom:} Interactive dependency graph visualization for a declaration (here, \texttt{String}), showing dependents up to configurable depth.
%     }
%     \label{fig:streamlit-explorer}
% \end{figure}

Overall, these experiments illustrate that \proofwala\ is not merely a proof search system, but a unified research framework for neural-guided theorem proving. Beyond reporting pass-at-$k$ metrics, our framework exposes assistant-level execution traces, maintains explicit environment state across parallel search branches, and logs fully annotated proof trees containing only compilable transitions. 

By combining multi-assistant interaction (\lean\; and \coq), environment pooling for scalable parallel search, and declaration-level metadata extraction, \proofwala\ enables detailed quantitative and qualitative analyses of search behavior. 
This includes measuring tree size, branching factors, proof multiplicity, execution latency, and structural transfer across assistants—analyses that are difficult to obtain from aggregate success rates alone. Such instrumentation transforms proof search from a black-box evaluation into a reproducible and analyzable experimental pipeline for studying neural-guided reasoning in interactive theorem provers.

\section{Related Work}
Previous open-sourced tooling has been developed for interaction with formal proof assistants, but individually only using a single language. Oftentimes, this tooling also contains data extraction features, compiling proof datasets from popular formalization repositories such as Mathlib \cite{mathlib} for \lean, CompCert \cite{leroy2009formal}, and Mathcomp \cite{githubGitHubMathcompmathcomp} for \coq. LeanDojo \cite{yang2023leandojo} provided open-source tooling for interaction with \lean~3 and extracted a proof step dataset from Mathlib.\footnote{LeanDojo now supports \lean~4, which is not backwards compatible to \lean~3.} NTP Toolkit \cite{ntptoolkit} supports extracting training data from arbitrary \lean~ repositories. CoqGym \cite{yang2019learning} is a framework for interaction and data collection with \coq\; up to versions 8.12.0 (because of dependency on SerAPI library\footnote{\href{https://github.com/rocq-archive/coq-serapi}{https://github.com/rocq-archive/coq-serapi}}). Proverbot \cite{sanchez2020generating} introduced Coq-Serapy, an interaction tool in \coq from which our \coq support is derived. CoqPyt \cite{CoqPyt} is a framework for interaction and data generation from \coq with emphasis on supporting LM-based methods. COPRA \cite{thakur2024incontext} introduces a framework for interaction with \lean~3 and \coq, but without tooling for data extraction or support for heavy parallelism during proof search. Aniva et al.~\cite{pantograph} introduced \pantograph, a \lean~4 interaction and proof tracing framework that supports tactic execution and proof-level data extraction. \pantograph\ enables collecting proof traces from \lean\ repositories and provides tooling for replaying and instrumenting proofs within \lean~4.

However, \pantograph\ is restricted to \lean\ and focuses primarily on tactic-level tracing. It does not provide cross-assistant abstraction, unified state representations, or repository-wide dependency analysis at the declaration level. In particular, it does not support extracting global dependency graphs, cross-file structural relationships, or multi-hop declaration dependencies in a form suitable for large-scale structural analysis or assistant-agnostic training.

In contrast, \proofwala\ is designed as a unified, cross-assistant framework supporting both \lean\; and \coq. Beyond proof-step extraction, it enables declaration-level dependency graph construction, SQL-based repository exploration, standardized multilingual dataset generation, neural model training pipelines, and scalable parallel proof search via environment pooling. This architectural unification allows controlled cross-lingual transfer experiments and fine-grained structural analyses that extend beyond tactic replay or single-assistant tracing.

A number of proof search methodologies have been proposed in the recent literature. GPT-$f$ \cite{polu2020generative} employed a best-first search approach with a trained transformer-based architecture for proof synthesis. LeanDojo \cite{yang2023leandojo} similarly employs a best-first search, though augments the neural prediction model with a retrieval model which predicts relevant premises. HyperTree Proof Search and ABEL \cite{lample2022hypertree, gloeckle2024abel} introduces an online variant of Monte Carlo Tree Search for the theorem-proving task. PACT \cite{han2021proof} introduces auxiliary training objectives derived from proof state data to learn a better prediction model for search. COPRA \cite{thakur2024incontext} uses large LMs as proof step prediction models, which can be conditioned on additional information such as retrieved lemmas, definitions, and execution information, for search. Graph2Tac \cite{blaauwbroek2024graph2taconlinerepresentationlearning} learns online hierarchical representations of definitions and theorems, and is used for proof search in Tactician \cite{Blaauwbroek_2020}. Several tools have been developed to help with live formalization efforts; these include LLMStep and LeanCopilot for \lean \cite{welleck2023llmstepllmproofstepsuggestions, song2024largelanguagemodelscopilots}, and CoqPilot for \coq \cite{coqpilot}.

Previous work has explored providing effective support for measuring models across various interactive theorem provers. \texttt{miniF2F} \cite{zheng2021minif2f} is a multi-language benchmark of high-school competition math problems formalized in \lean~3, HOL Light, Isabelle, and Metamath, though not in \coq. PutnamBench \cite{tsoukalas2024putnambenchevaluatingneuraltheoremprovers} is a collegiate-level benchmark for competition math in \lean~4, \coq, and Isabelle. We do not include evaluations on PutnamBench as our work is not targeted towards olympiad-style theorem-proving. MMA \cite{jiang2023multilingualmathematicalautoformalization} demonstrates that models trained on data from both languages yield downstream performance improvements for autoformalization in both languages, compared to models trained on just one language of data. In our experiments, we demonstrate that such transfer also occurs for neural models trained to perform proof step prediction.

\section{Conclusion}
We introduced \proofwala, a unified research framework for neural-guided theorem proving across interactive proof assistants such as \lean\; and \coq. The framework standardizes assistant interaction, proof-state extraction, dataset construction, model training, and scalable parallel proof search under a common abstraction. 

Using this infrastructure, we constructed a multilingual proof-step dataset spanning mathematics and software verification repositories and trained the first multi-assistant proof-step prediction models across \lean\; and \coq. Our experiments demonstrate that multilingual training provides evidence of transfer across assistants and domains. We observe the strongest improvements on the largest benchmark and in the domain adaptation setting. At the same time, other datasets show consistent, albeit smaller, gains over monolingual baselines in both proof-step prediction and end-to-end proof search. This suggests that the benefits of multilingual training are most pronounced in high-data regimes and when adapting to new domains.
%Our experiments demonstrate that multilingual training improves transfer across assistants and domains, outperforming monolingual baselines in both proof-step prediction and end-to-end proof search.

Beyond the empirical results, the primary contribution of this work is architectural. By unifying execution, data collection, dependency analysis, environment pooling, and search under a single framework, \proofwala\ enables systematic, reproducible, and fine-grained analysis of neural proof search across multiple ITPs. We view this unification as an essential step toward scalable and comparable research in automated theorem proving.

In future work, we aim to extend the framework to additional assistants and richer dependency representations, explore adaptive and learning-based search strategies within the unified abstraction, and investigate reinforcement learning approaches that leverage the logged proof-tree structures produced during search. We believe that such framework-centered development will be crucial for advancing neural-assisted formal reasoning.

\bibliography{references}

\newpage

\end{document}